\documentclass[10pt,twocolumn,letterpaper]{article}

\usepackage{cvpr}
\usepackage{times}
\usepackage{epsfig}
\usepackage{graphicx}
\usepackage{amsmath}
\usepackage{amssymb}
\usepackage[normalem]{ulem}

\usepackage[utf8]{inputenc}
\usepackage{listings}
\usepackage{float}
\usepackage{graphics}
\usepackage{graphicx} 
\usepackage{caption} 
\usepackage{subcaption} 
\usepackage{verbatim}
\usepackage{cite}

\usepackage[pagebackref=true,breaklinks=true,letterpaper=true,colorlinks,bookmarks=false]{hyperref}

\cvprfinalcopy 

\def\cvprPaperID{3265} 

\newcommand{\tompson}[1]{{\color{green} JT: #1}}

\newcommand{\edit}[1]{{\color{red} E: #1}}

\ifcvprfinal\fi
\begin{document}
\title{Beyond Photo Realism for Domain Adaptation from Synthetic Data}

\author{Kristofer Schlachter\\
NYU\\
{\tt\small schlacht@cims.nyu.edu}
\and
Connor DeFanti\\
NYU\\
{\tt\small cdefanti@cims.nyu.edu}
\and
Sebastian Herscher\\
NYU\\
{\tt\small herscher@nyu.edu}
\and
Ken Perlin\\
NYU\\
{\tt\small perlin@nyu.edu}
\and
Jonathan Tompson\\
Google Brain\\
{\tt\small tompson@google.com}
}

\maketitle
\maketitle

\section{Abstract}

As synthetic imagery is used more frequently in training deep models, it is important to understand how different synthesis techniques impact the performance of such models. In this work, we perform a thorough evaluation of the effectiveness of several different synthesis techniques and their impact on the complexity of classifier domain adaptation to the ``real" underlying data distribution that they seek to replicate. In addition, we propose a novel learned synthesis technique to better train classifier models than state-of-the-art offline graphical methods, while using significantly less computational resources.  We accomplish this by learning a generative model to perform shading of synthetic geometry conditioned on a ``g-buffer'' representation of the scene to render, as well as a low sample Monte Carlo rendered image. The major contributions are (i) a dataset that allows comparison of real and synthetic versions of the same scene, (ii) an augmented data representation that boosts the stability of learning and improves the datasets accuracy, (iii) three different partially differentiable rendering techniques where lighting, denoising and shading are learned, and (iv) we improve a state of the art generative adversarial network (GAN) approach by using an ensemble of trained models to generate datasets that approach the performance of training on real data and surpass the performance of the full global illumination rendering.
\section{Introduction}

Applying deep learning to supervised computer vision tasks commonly requires large labeled datasets \cite{imagenet, coco}, which can be time consuming, expensive or impractical to collect. As a result, increasingly more synthetic data has been used to overcome these scalability limitations \cite{DBLP:journals/corr/RichterVRK16, DBLP:journals/corr/ShafaeiLS16, DBLP:journals/corr/ZhangSYSLJF16, DBLP:journals/corr/SixtWL17}. However, using synthetic data can introduce problems that arise from the differences in the distributions of the real and synthesized data domains\cite{2014arXiv1409.7495G}. Usually the best way to reduce the differences is to use the most realistic simulation method available for synthesizing images, in this case a Global Illumination (GI) based renderer such as Mitsuba\cite{Mitsuba}.  Unfortunately, the computational cost of most GI renderers can be prohibitive.  Recently, there has been research into boosting the effectiveness of existing synthetic data\cite{DBLP:journals/corr/ShrivastavaPTSW16}.  However, these new techniques were not benchmarked against real RGB data or against other rendering techniques.

This work explores the effectiveness of various rendering methods, which range from very low to very high computational cost. We compare multiple synthetic rendering methods with the performance obtained using the "real" data distribution as an optimal baseline. This is made possible since we annotate an existing image classification dataset (NORB \cite{LeCun:2004:LMG:1896300.1896315}) with pixel aligned geometry labels obtained using a high-quality photometric scanner. We represent the geometry and material information of the RGB scene using a "Geometry Buffer" or "g-buffer", which gives us a standard set of 3D scenes where geometric transformations are frozen in place. By using a g-buffer, we isolate the physical process of shading a surface from the geometric process of geometry pose and projection.  The shading process is also decoupled from camera parameters, including location, orientation and field of view. This allows us to directly compare different rendering methods and enables us to measure any performance difference due to shading alone.

Capitalizing on this standard 3D scene representation, we setup a number of experiments to compare rendering methods of increasing sophistication and computational cost.  We then measure the benefit of adding these rendering techniques and see if the computational cost can be justified.  These experiments measure the relative importance of image rendering features. We use a full GI rendering of Mitsuba\cite{Mitsuba} as the method whose performance we sought to match or surpass.
We are able to do this by using non-standard rendering techniques that use convolutional networks to learn a domain optimal shading function. We use this network as a form of rendering where the shading is differentiable. Effectively, we use learned shading to denoise low sample count GI rendering and compared it to high sample count images.  We also explore the use of Generative Adversarial Networks to learn shading functions that could render images in unique ways that even outperform the accurate, but extremely expensive GI output.
\subsection{Contributions}
\begin{enumerate}
\item We provide a dataset (which we will make public) that allows the direct comparison of training on real data versus synthetic by including 3D scans of the objects that are in the photographic dataset.
\item We introduce a novel method of using an ensemble of GAN trained generative models to create an augmented dataset.
\item We contribute a thorough comparison of the performance of all the rendering methods with real data.
\end{enumerate}
\section{Related Work}

Previous papers have used g-buffers to condition a learned generative model for shading. Nalbach et al. \cite{DBLP:journals/corr/NalbachAMSR16} uses a similar model architecture as the denoiser we employ as well as using a g-buffer as input.  The authors use it to learn to approximate screen space shading techniques with a data-driven approach, to see if data driven methods can replicate the standard shading methods used in games. They differ in their approach in that they are optimizing for images looking real to humans and not optimizing the images' utility for training classifiers or to improve domain adaptation.


There have been recent advances in denoising low sample count images calculated by Global Illumination (GI) based renderers.
Shied et al. \cite{Schied:2017:SVF:3105762.3105770} denoises low sample count GI images by combining neighboring frames and feeding them into a fixed spatio-temporal filter.  The authors emphasis is on temporal stability, efficient computation and denoising single sample frames in real-time. Their authors do not employ machine learning in their method. 
Chaitanya et al. \cite{Chaitanya:2017:IRM:3072959.3073601} along with 
Bako et al. \cite{Bako17} and 
Kalantari et al. \cite{Kalantari:2015:MLA:2809654.2766977} all use machine learning to denoise Monte Carlo (GI) renderings for graphics applications.  This work uses a modified form of the model presented in \cite{Chaitanya:2017:IRM:3072959.3073601} in order to denoise images used for training classifiers with the goal of synthesizing training data faster.

Further use of global illumination for training classifiers has been explored.
Zhang et al. \cite{DBLP:journals/corr/ZhangSYSLJF16} use a custom created large 3D indoor dataset to conduct experiments relating the performance of different renderers and lighting for various machine learning tasks, such as normal estimation, semantic segmentation, and object boundary detection. The approach presented in this work, in contrast, focuses mainly on shading and its effect on training data quality.



Using a video game engine for dataset creation has been explored by Richter et al. \cite{DBLP:journals/corr/RichterVRK16} and  Shafaei et al. \cite{DBLP:journals/corr/ShafaeiLS16}. They explored the use of video game snapshots to show that it could train image segmentation models. We build upon their approach by using learned rendering techniques to improve the synthesis quality and to decrease rendering computational cost.


To learn shading, we used Generative Adversarial Networks (GANs) which were first introduced by Goodfellow et al. \cite{goodfellow} to learn complex multi-modal distributions. Since then many works have used them to synthesize images. 


Learning, partially differentiable renderers was proposed by Sixt et al. \cite{DBLP:journals/corr/SixtWL17}. In their paper they train a differentiable 3D renderer, except their model is scene and application specific. The differentiable rendering model architecture we implemented is not restricted to a specific shading model, geometric input or lighting condition.

A GAN based method that uses unlabeled data to refine realistic rendered images was pioneered by
Shrivastava et al. \cite{DBLP:journals/corr/ShrivastavaPTSW16}. The paper proposes an architecture that takes in rendered images and makes them more realistic by using an adversarial loss to tune a refiner model. We use this paper as a starting point for our GAN experiments; however, we modify many aspects of it to suit our input, which is a g-buffer (not a pre-rendered image).

A GAN-based method related to Shrivastava et al. is Bousmalis et al. \cite{Bousmalis2016UnsupervisedPD}. This paper proposes a very similar method, but conditions the input with a noise vector and a depth buffer.  The depth buffer is used to mask out the background so the refiner network can concentrate upon pixels containing the object to be classified.  Our method conditions on more data including normals and albedo and does not mask out parts of the image which allows for shadowing to be modeled. We also improve upon these two papers by using an ensemble of trained GAN generative models to boost training data performance.



\subsection{Rendering Methods and Techniques}

\subsubsection{G-Buffer}

\begin{figure}[h!]
\centering
\includegraphics[width=0.8\columnwidth]{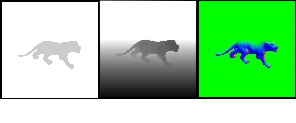}
\caption{G-Buffer. Left is Albedo, middle is Depth, right is world space Normals.}
\label{fig:GBUFFER_ALBEDO} 
\end{figure}

G-Buffers \cite{Saito:1990:CRS:97879.97901} separate the geometric processes such as transformation, projection and depth testing, from physical processes such as shading and texture mapping. This is accomplished by the buffers accumulating the geometry information in the first pass and then lighting each pixel of the output on a second pass (a process that is linear in both model and lighting complexity).  This is in comparison to a traditional ``forward renderer'' which calculates geometry information and performs lighting and shadowing in one pass, followed by and writing the result to the output buffer (with quadratic complexity). The technique of using G-Buffers and screen-space rendering is known as ``deferred shading,'' where expensive lighting and shadowing computation is done after all of the visible geometry is encoded into its buffers.
 As an added benefit for this work, the encoding of the geometry state into G-Buffers allows for simplified testing of different shading techniques on the identical 3D scenes.
In this work's experiments, the G-Buffer consists of the scene encoded geometry normals, geometry diffuse color (also referred to as albedo), and the linear depth from the camera to the visible geometry.  See figure \ref{fig:GBUFFER_ALBEDO}, for examples. We describe how the datasets are produced in \ref{TrainingData}.

\subsubsection{Spherical Harmonics}
\begin{figure}[h!]
\centering
\includegraphics[width=1.0\columnwidth]{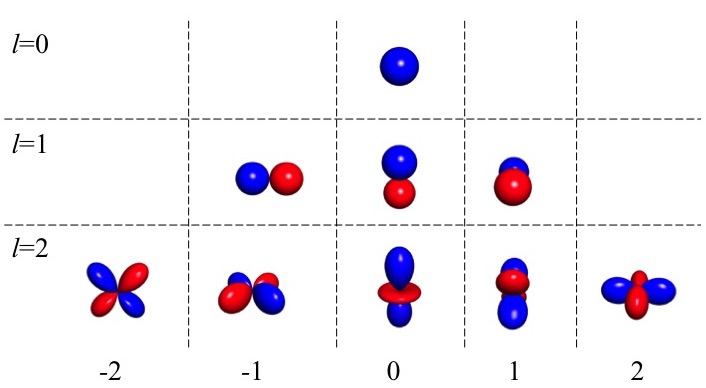}
\caption{Visualization of first 3 bands of the spherical harmonics functions. Red are negative values and blue are positive. These are the corresponding 9 functions to the 9 coefficients we learn for lighting.}
\label{fig:SHBands01}
\end{figure}

In this work's experiments, we use a set of functions called Spherical Harmonics (SH) to learn the lighting of an environment. Spherical Harmonics are a set of continuous and rotationally invariant functions that approximate an arbitrary spherical function, such as the distribution of light reflected off a surface \cite{Ramamoorthi:2001:ERI:383259.383317}, also known as a bidirectional reflectance distribution function (BRDF). The orthogonal functions are grouped into levels of increasing detail, resolution, or frequency. SH are grouped into levels, where each level is a set of orthogonal functions of finer detail, resolution, or frequency \cite{Shreiner:2013:OPG:2544032}.  To represent the smooth lighting, we only need the first three levels which are a constant, linear polynomials, and quadratic polynomials of the surface normal. See Figure \ref{fig:SHBands01} for a visualization of the first three levels.


An advantage to using Spherical Harmonics for lighting is that you can represent a smooth lighting environment with just 9 floats. This allows us to parameterize a lighting function with far fewer parameters compared to other representations. This has the effect of drastically reducing the search space when trying to learn a parameterized lighting model.

We define a trainable module that calculates three levels of SH basis functions. The input to the module consists of the surface normal at every pixel, which is conveniently stored in the G-Buffer. The network outputs a shaded image which can be composited with shadowing and albedo.

\subsubsection{Ambient Occlusion}

In an environment with only large area light sources, there exists a simple first order approximation to global illumination called Ambient Occlusion. It is used as a form of simplified global illumination because it takes into account only screen-space visible scene geometry\cite{Miller:1994:EAL:192161.192244}.  Each view-ray from the camera's pixels defines a single point on the surface of the scene geometry. The algorithm uses these local screen-space geometry samples to approximate how exposed each surface fragment is to incoming ambient light sources by estimating how much of the hemisphere around that point is blocked by nearby surfaces. It is an approximation since the geometry visible to the camera may not represent all local geometry around the surface patch. However, for lighting conditions that are close to a uniformly bright hemisphere, like an overcast day, and for simple geometry, then ambient occlusion is a good approximation of global illumination.

\subsubsection{Global Illumination}\label{mitsuba_section}

Global illumination (GI) refers to rendering methods that take into account inter-surface reflection, to significantly improved visual fidelity in comparison to faster approximate methods that only simulate direct lighting (or limited bounce counts).  GI derives its accuracy from its ability to take into account the light that arrives at a surface not just directly from a light source, but from light that has been reflected off of other surfaces as well. Different surface materials have different reflection characteristics that effect how much sampling must be done in order to converge to the correct rendering solution. Each dataset required different sample counts to converge due to the different material distributions.


\section{Training Data Creation}\label{TrainingData}
Direct comparisons between rendering techniques and real images could not be done without creating a dataset that has approximately 1:1 correspondence between domains.  To this end we identified the NYU Object Recognition Benchmark (NORB) dataset \cite{LeCun:2004:LMG:1896300.1896315} as a good starting point. It is small enough to allow for recreation of each photograph with high accuracy in simulation.  A characteristic that makes NORB even more attractive is that the object location, rotation, camera placement and lighting condition are known for every image. Furthermore, each object was modified so that the BRDF of each object was approximately homogeneous across the dataset.

However, the NORB dataset is comprised of grayscale images, which is less-relevant for modern deep-learning architectures. As such, we composed an additional dataset of synthetic objects from ShapeNet~\cite{DBLP:journals/corr/ChangFGHHLSSSSX15} (choosing a disjoint set of object classes), where the ground-truth RGB image is the result of our most expensive offline rendering method available. This dataset was digital-only and did not have physical models that we could scan like the NORB dataset. So, we used synthetic renders for both training the learning model and testing its accuracy.

While this is not sufficient on its own to make conclusions about the learning model used in this work on real image data, we use this additional synthetic dataset to validate the grayscale NORB results on a RGB dataset of similar complexity and composition.

\subsection{NORB Dataset}\label{NORB-DATASET}
\begin{figure}[h!]
\centering
\includegraphics[width=1.0\columnwidth]{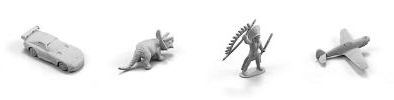}
\caption{A sample of the NORB test dataset. Note the dark shadows under some of the models, animals in particular.}
\label{fig:norb-samples}
\end{figure}

The NORB dataset consists of a set of 50 toys split into a 25 object training set and a 25 object test set. For each object, stereo photographs were taken with the object on a turntable in 6 controlled lighting conditions and with 18 camera azimuth angles and 9 elevation angles. This results in a dataset with 48,600 stereo pictures.  We limited the dataset to just the 25 objects in the training set. We also confined the dataset to one lighting condition, reducing the image count in the training dataset to 4,050 images. The toys were painted a uniform color with a matte paint. This was intended to prevent texture from being used in classification.

 All of the properties listed above facilitate a near-perfect recreation of the original scenes through 3D rendering (whereas it would be significantly more complicated to recreate natural outdoor scenes with perfect 1:1 geometry correspondence). We know the pose, location and surface of the objects as well as the location of the camera. We set the material properties based on the paint used for the physical toy models.
 
We used the HP 3D Structured Light Scanner Pro S3\cite{HPSCANNER} to scan in the objects. The models were then uniformly scaled to fit into a unit cube with the bottom center at the origin. We then rendered a matching g-buffer for every image in the training dataset.  

Mitsuba produced the G-Buffers and shaded images at a resolution of 768x768 for each image. In order to compare results based on image quality, we generated 128, 4, and 1 sample count images, which are later referred to as the high, medium, and low sample count images. The images and buffers were then cropped and resized such that the bounding box of the mesh pixels fit into an 80x80 pixel square, with the final image itself being a 96x96 pixel square. This mirrors the procedure done for the source NORB dataset. This creation procedure results in a one-to-one mapping between each image in the synthetic training dataset with the corresponding ``real'' image in the NORB training dataset.
\subsection{ShapeNet Dataset}

\begin{figure}[h!]
\centering
\includegraphics[width=0.24\columnwidth]{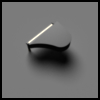}
\includegraphics[width=0.24\columnwidth]{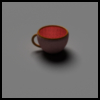}
\includegraphics[width=0.24\columnwidth]{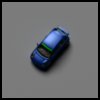}
\includegraphics[width=0.24\columnwidth]{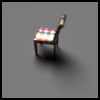}
\caption{A sample of the ShapeNet train dataset.}
\label{fig:SHAPENET}
\end{figure}

  The ShapeNet dataset consists of tens of thousands of models. We use a subset of ShapeNet: 10 different models from 10 categories each. The dataset size was chosen to match similar popular, small datasets CFAR-10 \cite{Krizhevsky09learningmultiple} and MNIST \cite{726791} and since we want to measure the relative performance of our domain adaptation techniques in the regime of limited training data (which more closely matches an intended use case). The categories we chose were airplanes, boats, cars, chairs, motorcycles, mugs, pianos, planters, tables, and trains. This provided a wide variety of models that could be identified from many different camera angles. 

After each model was changed to fit the above conditions, they were then rendered using the Mitsuba renderer. An infinitely large, matte ground plane was created below the objects, and two large, spherical area lights were created above the object to create shadows. In a specific experiment, detailed later, these lights were replaced by a single directional light to create harder shadows. For each model, we used 162 randomly sampled camera angles. The camera angles were chosen such that the elevation angles were between 30 and 70 degrees, and the azimuth could be any angle (between 0 and 2$\pi$ radians). These bounds and the number of samples were chosen to best match the NORB dataset. The randomization was chosen uniformly, such that if $\zeta_1\in[\cos(70^\circ),\cos(30^\circ)],\zeta_2\in[0,1]$ are uniformly distributed random numbers, then the elevation angle $\theta$ and azimuth angle $\phi$ are $\theta = \cos^{-1}\zeta_1$, $\phi = 2\pi \zeta_2$. This follows the formula given by \cite{Pharr:2010:PBR:1854996}. The images were rendered at 96x96, outputting the G-buffer of each image, including the image albedo, normals, occlusion, distance field, and GI image data. For this GI rendered dataset, we generated 1024 (high), 32 (medium), and 10 (low) sample count images.

\section{Experiments}

In the following experiments, we use increasingly complex simulation methods for rendering the images.  We start with just a silhouette as training data and we then incrementally increase the sophistication level of the rendering.  We then add complexity such as shading, shadowing and global illumination. Finally, we use a form of a denoising autoencoder to learn shading, cleanup and refine GI based images.

\begin{figure}[h!]
\centering
\includegraphics[width=1.0\columnwidth]{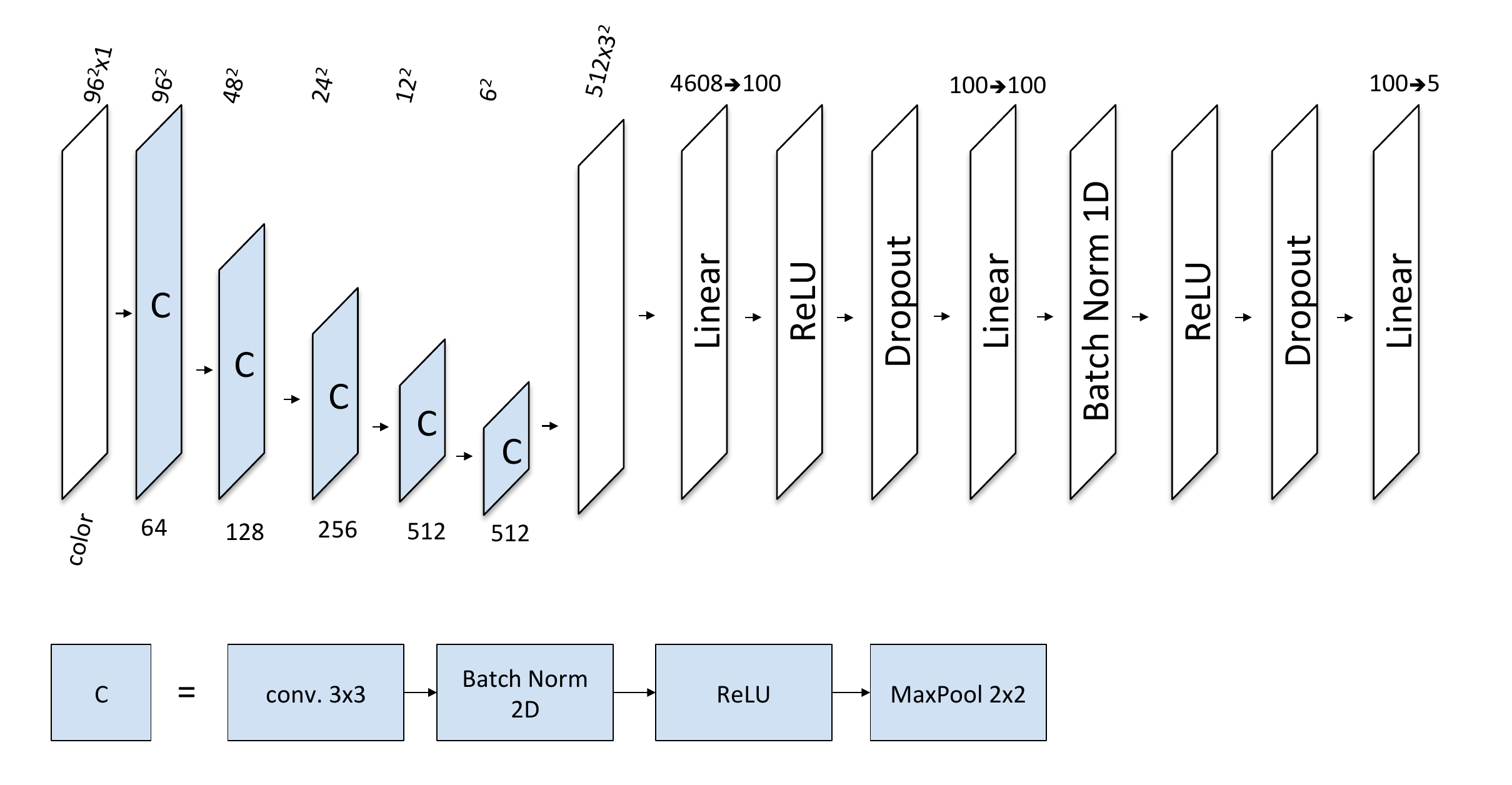}
\caption{Architecture of VGG08 Batch Norm Classifier}
\label{fig:VGG_DAIGRAM}
\end{figure}

In order to principally measure the impact of these various techniques, we use a single classifier architecture across all experiments; a simplification of the VGG network\cite{DBLP:journals/corr/SimonyanZ14a}. Given the small size of the datasets, we reduce the standard VGG network to 8 learnable layers, including 5 convolutional and 3 fully-connected layers along with a cross-entropy loss (negative log likelihood). See Figure~\ref{fig:VGG_DAIGRAM} for an overview of this architecture.

The first experiment done for a dataset is to train a classifier with data that is from the same domain as the test set.  In NORB, this is using the real photographs for training and in ShapeNet, it is the highest sample count Mitsuba rendered images. 

This establishes a target classifier accuracy to be used for measuring the accuracy from using the training data created by the various rendering methods below. Next, we conduct multiple experiments by varying rendering techniques and measuring their quantitative impact on classifier performance.

Firstly, we use renders that use \textit{no shading} by taking the albedo maps of each image generated in the g-buffer and using this to train the dataset. We then test this against the baseline generated in the first experiment and compare the results. The output images can be seen in figure \ref{fig:GBUFFER_ALBEDO}.

\begin{figure}[h!]
\centering
\includegraphics[width=1.0\columnwidth]{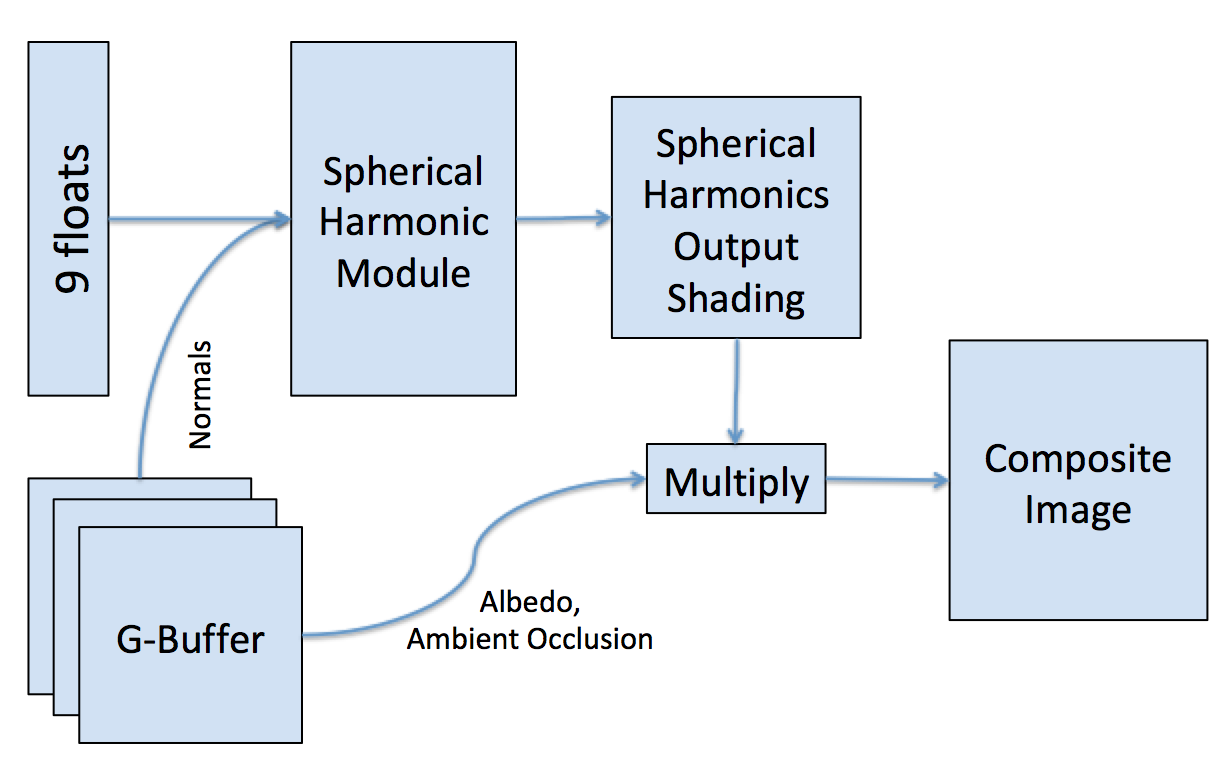}
\caption{Spherical Harmonics network architecture.}
\label{fig:SHN}
\end{figure}

\begin{figure}[h!]
\centering
\includegraphics[width=1.0\columnwidth]{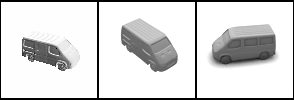}
\caption{Left: Learned SH. Middle: SH x Albedo. Right: SH x Albedo x Ambient Occlusion.}
\label{fig:SHComparison}
\end{figure}
The next step beyond albedo-only rendering is shading the models in the image, but not rendering shadows. In order to do this, we learn the shading using the learnable spherical harmonics module (the advantage of learning the diffuse shader is that we can more closely approximate the statistics of the ``real'' image domain), as seen in figure \ref{fig:SHN}.  We then take these simple shaded renderings and composite them with ambient occlusion (AO) present in the g-buffer as an approximation of lighting condition independent shadowing. An example output image can be seen in figure \ref{fig:SHComparison}. As this is the first experiment that uses shadows, this experiment determines whether or not shadows are important in image classification.

\begin{figure}[h!]
\centering
\includegraphics[width=1.0\columnwidth]{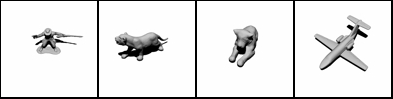}
\caption{Two Bounce Rendering.  Note that it is the equivalent to a shadow mapped scene and differs from the AO composited scenes by the sharpness of the shadows. This is due to the scene using a single directional light.}
\label{fig:2BOUNCE}
\end{figure}
While AO makes a good, fast approximation of shadows, they are not very realistic. Therefore, we next test more realistic shadows. We wish to measure the impact of inter-reflections and global illumination in a later experiment, and so we limit the renderer to do 2-bounce raytracing with a directional light. This allows us to create very hard shadows, as seen in figure \ref{fig:2BOUNCE}.

\begin{figure}[h!]
\centering
\includegraphics[width=1.0\columnwidth]{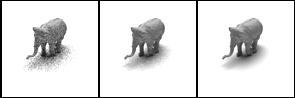}
\caption{
Mitsuba renders at different sample counts. Left: 1 sample per pixel.  Middle: 4.  Right: 128.}
\label{fig:differentsamplesraw}
\end{figure}
Finally, we use a full (GI) renderer to simulate a realistic scene. Using Mitsuba, this is the closest we can get to the baseline data. As noted earlier, Mitsuba is a Monte-Carlo raytrace-based renderer, and therefore allowing for more samples-per-pixel will produce a more realistic image. However, as a drawback, higher sample counts take linearly more time to render (on the order of seconds for low sample count data and minutes for the high sample count data). We test on three different sample counts for each of the datasets to determine how much is gained from using higher values. The difference in sample counts can be seen in figure \ref{fig:differentsamplesraw}.

\begin{figure}[h!]
\centering
\includegraphics[width=1.0\columnwidth]{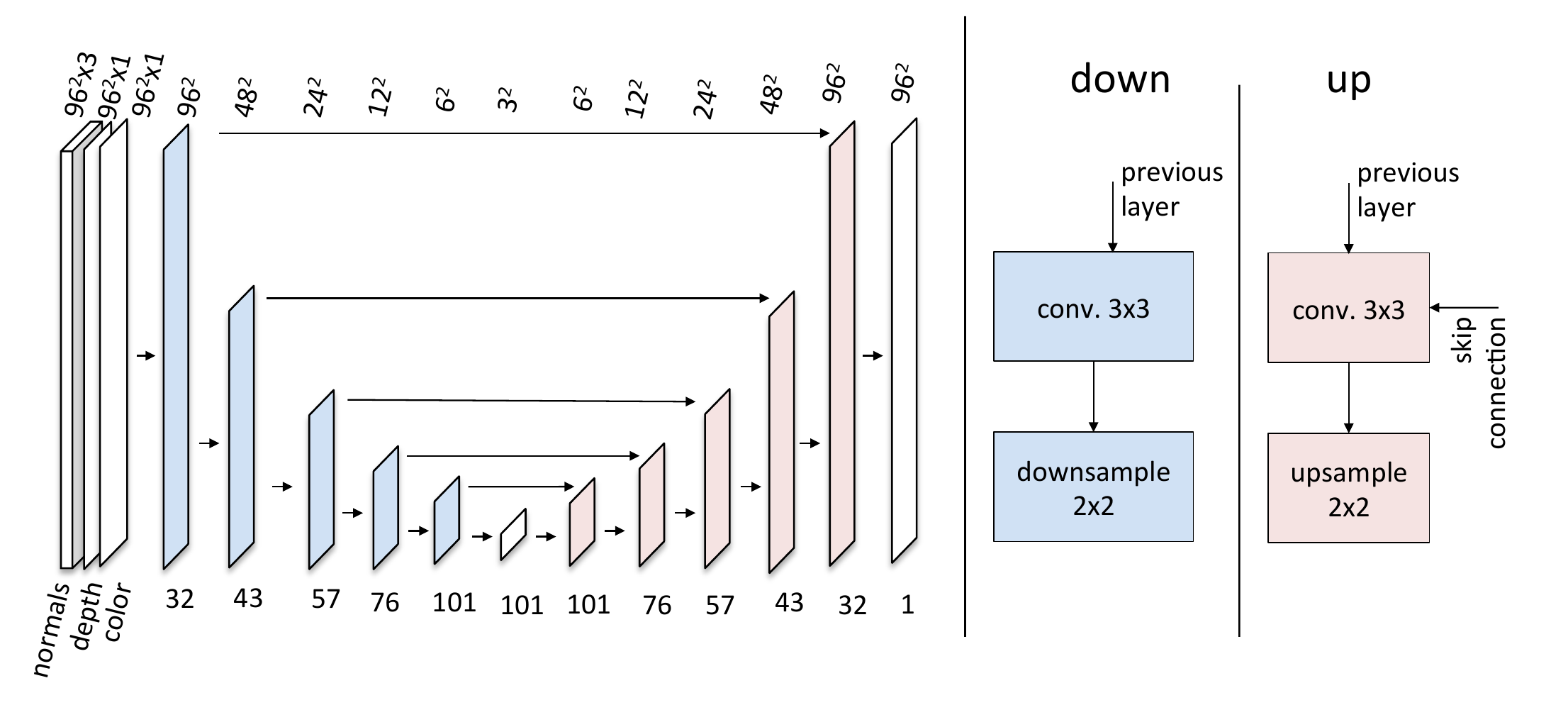}
\caption{Architecture of Denoising Model}
\label{fig:denoise}
\end{figure}
Given that the low sample GI renders take far less time than the high sample renders, we next experiment to see if the lower sample count images can be denoised using a simplified version of the denoising auto-encoder with skip connections inspired by \cite{Chaitanya:2017:IRM:3072959.3073601}. 
The network architecture is shown in \ref{fig:denoise}. For this network, we use the Structured Similarity (SSIM) loss function \cite{Wang04imagequality}, a loss function that has proven to be useful for both image-compression measurement and also image comparison. 

This denoising process takes milliseconds on the modern GPU, which is a far smaller cost than rendering the high-sample GI renders.


\subsection{Train on GAN Images}

Another synthetic data generation technique that we explored involved the use of Generative Adversarial Networks (GANs) \cite{2014arXiv1409.7495G}.  We use a GAN architecture inspired by \cite{DBLP:journals/corr/ShrivastavaPTSW16} and \cite{Bousmalis2016UnsupervisedPD}. This work differs from \cite{Bousmalis2016UnsupervisedPD} and \cite{DBLP:journals/corr/ShrivastavaPTSW16} by conditioning on a g-buffer and rendered images with and without the presence of sampling noise. Furthermore, this work doesn't require masking the input images or using local loss functions. This is due to an interesting property of the rendering architecture whereby conditioning the generative model on a static g-buffer, and by ensuring that the shading function is dependent on geometry normals (which are not modified by the generative model and is physically motivated by the rendering equation) 
prevents the need for a local loss function and eliminates the problem of a GAN altering the semantic content (i.e. labels) of the objects being rendered.
The GAN architecture we employed was the following:  
The generative network was the same model used for denoising shown in figure \ref{fig:denoise}.  The inputs were normals, depth and either the low, medium and high sample global illumination rendered image. Albedo was used for only ShapeNet data based experiments. The regularization loss was SSIM.

Pre-training the generative network against the high sample Mitsuba rendering before starting adversarial training produced a partially differentiable renderer that learned to denoise and shade the scene conditioned on the g-buffer and image input. Unlike in \cite{DBLP:journals/corr/ShrivastavaPTSW16}, the conditional data distribution in this work is visually very similar to the target generated distribution. As such, we propose a more principled way to validate the performance of the GAN during training (the proposed method in \cite{DBLP:journals/corr/ShrivastavaPTSW16} was to ``visually inspect'' the GAN output and early-stop training when the output visually resembled the target dataset). To choose a GAN to generate training data we train a classifier on real training and validation data.  No test data is used in testing or training this classifier.  We will call this a clean classifier. For every saved GAN model, we run the synthetic training data through it and establish how well the clean classifier can classify the refined images. We use the classification accuracy to rank each GAN. We then take the top 10 GAN refiner models that were ranked from the clean classifier and for each one train a classifier on the full dataset. For an ensemble of GAN models (described below) we take the set of top 10 ranked refiners from the clean classifier and train a classifier with a subset of the refiners synthesizing a new expanded dataset.

The expanded dataset is created by the following method. Firstly, we create a new empty training set. For each generative model in the set of refiners we refine each image in the original full training set and append it to the new training set. The new dataset is comprised of $D * N$ images where $N$ is the number of refiners and $D$ is the dataset size.
This effectively generates $N$ copies of the training data where each copy is unique and without the computational overhead of a full render to synthesize the images.

Taking the top 10 models ranked by this method, some models outperform just using a raw low sample image or the result of denoising it.  This method of choosing a GAN allows for selecting a model that outputs non-photorealistic images and represents a principled way to choose models independent from the realism of their generated input.  When evaluating performance of using a single GAN, we took each saved GAN and trained a classifier.  We repeated this three times and then kept the top performing model. For every sample count we found a GAN that could outperform directly training on the original image or denoising it.


\begin{figure}[h!]
\centering
\includegraphics[width=1.0\columnwidth]{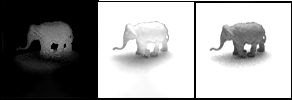}
\caption{Right: 4 sample Mitsuba image passed into GAN, Middle: Output of refiner.  Left: Absolute value difference between the full sample rendered image and the refined image. Notice that the image is no longer photo-realistic.}
\label{fig:GAN_4}
\end{figure}

\subsubsection{Training Using an Ensemble of GANs} \label{sec:multigans}
The next experiment is a little different from the rest, as more training data is used and is therefore not quite comparable.  However, the amount of training data that would have to be generated in a GI renderer remains the same.  The major change is the use of the top ten ranked ensemble of GAN models to create multiple refined version of the data.

The new dataset is an aggregation of the different refined datasets.  The costly pre-rendering of the input buffer and images is not changed and the run-time rendering refinement is longer but insignificant compared to GI rendering time.  In our experiments, it took a minute longer to generate 10 times as much data by running the dataset through ten GANS on an NVIDIA Titan X Maxwell GPU. 
The added time for using this method is negligible except for extra time required to train on a larger dataset. This ensemble of GANs is another method in this work that distinguishes it from \cite{Bousmalis2016UnsupervisedPD}.

\section{Results}
We now examine the results from our previous experiments. As the patterns in results between the NORB and ShapeNet datasets were very similar, we will first detail the NORB results, and then use the ShapeNet results to validate these findings on RGB.
\begin{table}[]
\centering
\begin{tabular}{|l|r|r|}
\hline
\multicolumn{1}{|c|}{\textbf{Input}}
& \multicolumn{1}{r|}{\textbf{Accuracy}}
& \multicolumn{1}{r|}{\textbf{Baseline \%}} \\ \hline
Albedo Only 				&53.26\%	& 56.06\%	\\
Learned SH Shading			&53.06\%	& 55.85\%	\\
SH Shading + AO				&60.89\%    & 64.09\%   \\
2 Bounce (Direct Lighting)	&66.49\%	& 69.98\%   \\
\textbf{Real Images}		& 95.01\%	& 100.00\%	\\ \hline
\multicolumn{3}{|c|}{\textbf{NORB Performance of Basic Rendering Methods}}	\\ \hline
\end{tabular}

\caption{Performance of real-time (rasterization-based) rendering methods on the NORB dataset. Accuracy is the classification accuracy of the method, and Baseline \% is how good the method was compared to the baseline render.}
\label{table:tblnonGI}
\end{table}
\subsection{NORB Results}
The results for the NORB dataset are shown in table \ref{table:tblnonGI}. Albedo Only (unshaded renders) yielded only 53.26\% accuracy. This tells us that while the shape is helpful in classifying an object (compare 53.26\% to 20\% from randomly guessing), it is not nearly enough information to be consistently accurate. This is also the case for the learned SH shading, which scored slightly under the unshaded renders at 53.06\%. We can see that in this case, the fake shading was no more helpful than no shading at all.

We observe slight improvements once shadows are added. AO, a first approximation to shadows, scored at 60.89\% accuracy, a significant jump from renders without shadows. Then, once we improved the quality of shadows with the 2-bounce rendering method, we saw another significant jump to 66.89\% accuracy. However, this is still far less than the target baseline.

\begin{table}[]
\centering
\begin{tabular}{|l|r|r|r|r|}
\hline
\multicolumn{1}{|c|}{\textbf{Input}}
& \multicolumn{1}{c|}{\textbf{Low}}
& \multicolumn{1}{c|}{\textbf{Medium}}
& \multicolumn{1}{c|}{\textbf{High}}
& \multicolumn{1}{c|}{\textbf{Baseline}} \\ \hline
Mitsuba& 64.44\%	& 73.13\%	& 74.74\%	& 78.67\% \\
Denoised& 71.48\%	& 73.56\%	& N/A 		& 77.42\%	\\
GAN& 72.84\%	& 77.65\% 	& 75.26\%	& 81.73\%	\\
Ensemble& 85.23\%	& 84.37\% 	& 87.33\% 	& 91.92\%		\\ \hline
\multicolumn{5}{|c|}{\textbf{Performance of GI Rendered Refined Images}}	\\ \hline
\end{tabular}
\caption{Comparison of GI Rendered Refined Image Performance. The GAN output refers to if only one GAN was used to refine the dataset or if the dataset was expanded by multiple GANs.}
\label{table:tblallrefined}
\end{table}

We next observe the results of the more advanced techniques, such as using GI rendering at a high sample count, denoising a low sample count GI rendering, and learning to render using GANs. These results can be observed in table \ref{table:tblallrefined}. While noisy (low sample) GI renders had poor performance, ranking close to the 2-bounce condition we observed earlier, the highest sample GI render achieved 74.74\% accuracy, almost 10\% higher. However, the medium sample count renders performed almost as well as the high sample count renders, demonstrating that beyond a certain quality threshold, the increase in accuracy becomes marginal. 

We can further close the gap between the high sample count renders and the lower sample count renders with the denoising auto-encoder, which brought the low sample count images to 71.48\% and the medium sample count images to 73.56\% accuracy, less than 1\% difference to the high sample count render. We can therefore conclude that images could be rendered much faster at low samples, processed with a denoising auto-encoder, and achieve very comparable classification accuracies with a much lower render time.

Finally, we can see that the images generated using the GANs detailed above had a better performance than any other result. With a single GAN, we achieve 75.26\% classification accuracy, which is only a slight increase over the best GI render results. However, with an ensemble of GANs, we were able to achieve an accuracy of 87.33\%, over 90\% of what one can achieve training with real photographic data. \textbf{A surprising detail concerning the GAN-based images is that the best performing generated images were not necessarily visually realistic as compared to the test dataset.} This result demonstrations that conventional measures of photorealism does not necessarily equate to improved domain adaptation performance via a Convolutional Network classifier. See Figure \ref{fig:GAN_4} where the refined images have exaggerated edges, shadows and reflective properties.
\subsection{ShapeNet Results}

\begin{table}[]
\centering
\begin{tabular}{|l|r|r|}
\hline
\multicolumn{1}{|c|}{\textbf{Input}}
& \multicolumn{1}{r|}{\textbf{Accuracy}}
& \multicolumn{1}{r|}{\textbf{Baseline \%}} \\ \hline
Albedo Only 				&44.44\%	& 52.91\%	\\
Learned SH Shading			&47.80\%	& 56.91\%	\\
2 Bounce (Direct Lighting)	&70.86\%	& 84.37\%   \\
\textbf{1024 Sample Renderings}		& 83.99\%	& 100.00\%	\\ \hline
\multicolumn{3}{|c|}{\textbf{ShapeNet Performance of Basic Rendering Methods}}	\\ \hline
\end{tabular}
\caption{Performance of real-time (rasterization-based) rendering methods on the ShapeNet dataset.}
\label{table:tblnonGI_sn}
\end{table}

\begin{table}[]
\centering
\begin{tabular}{|l|r|r|r|}
\hline
\multicolumn{1}{|c|}{\textbf{Input}}
& \multicolumn{1}{c|}{\textbf{Low}}
& \multicolumn{1}{c|}{\textbf{Medium}}
& \multicolumn{1}{c|}{\textbf{Baseline \%}} \\ \hline
Mitsuba		& 81.68\%	& 82.30\%	& 97.99\% \\
Denoised	& 78.06\%	& 79.12\%	& 96.25\%	\\
GAN	& 83.48\%	& 83.59\%	& \textbf{99.52\%}	\\
Ensemble& 86.48\%	& 86.65\% 	& \textbf{103.17\%}		\\ \hline
\multicolumn{4}{|c|}{\textbf{Performance of GI Rendered Refined Images}}	\\ \hline
\end{tabular}
\caption{Scores using the ShapeNet dataset. Note that the Multiple GAN method performs better than the highest sample render.
}
\label{table:tblallrefined_sn}
\end{table}

Our ShapeNet experiments resulted in similar performance characteristics as on the greyscale NORB dataset, as shown in table \ref{table:tblnonGI_sn}. We can see that with the unshaded or learned SH shading models the classification accuracies are very poor, approximately half of the baseline. Once we move to 2-bounce lighting with shadows, this becomes much closer. We can account for the relatively larger jump in performance when adding shadows compared to NORB because the lighting conditions between the 2-bounce render for ShapeNet were much more similar to the target renders than in the case with the NORB 2-bounce renders and their target. 

Similarly, due to the nearly identical lighting conditions, we can see in table \ref{table:tblallrefined_sn}, the lower sample renders performed nearly as well as the 1024 sample renders that were used as the baseline. Interestingly, the denoiser reduced classification accuracy. This suggests that the denoiser may have modified image statistics that were present in all sample count renders. 

However, where we see the best performance is when we render with GANs, much like NORB. With a single GAN, we achieve 83.59\% classification accuracy, or 99.52\% of the baseline. When we used an ensemble of GANs, we achieve 86.65\% accuracy, which is even more accurate than the baseline. This means that the GAN-rendered models are as good or better for image classification than the high sample GI renders and are an effective regularizer for classifier training.

\section{Conclusion}

By constructing a novel dataset consisting of synthetic and real images with extremely high levels of geometry correspondence, we are able to directly compare various image synthesis techniques and their impact on domain-adaptation. We do so by measuring the domain-transfer classification accuracy of a multi-class image classifier as a proxy measure for ``image-realism'' as perceived by a Convolutional Network classifier architecture.  We have shown that diffuse shading alone gives similar performance to a simple silhouette image, and that shadows are important for domain transfer.  Most importantly, we proposed two methods to approximate the effectiveness of offline global-illumination rendering, and whose domain adaptation performance actually exceeds it through the use of a learned shading function. The two learned methods optimized for image statistics that are not captured in standard rendering methods which do not take into account the discriminative capacity of convolutional networks. 
Furthermore, we have also shown that by using an ensemble of GAN models you can outperform GI simulation and  approach the performance of using real data.

\newpage

\bibliography{bibliography}{}
\bibliographystyle{plain}
\end{document}